\documentclass[conference]{IEEEtran}
\IEEEoverridecommandlockouts
\usepackage{cite}
\usepackage{booktabs}
\usepackage{amsmath,amssymb,amsfonts}
\usepackage{algorithmic}
\usepackage{graphicx}
\usepackage{textcomp}
\usepackage{xcolor}
\usepackage{xcolor}
\usepackage{stfloats}
\usepackage[most]{tcolorbox} 
\definecolor{boxgray}{HTML}{F9F9FB}   %
\definecolor{mblue}{HTML}{1A4C9C} 
\definecolor{titleblue}{HTML}{2E5FA3} %
\definecolor{frameblue}{HTML}{C6D6F2} %
\usepackage{caption}
\tcbset{
    colback=boxgray,
    colframe=frameblue,
    coltitle=titleblue,
    boxrule=0.5pt,
    arc=1mm,
    left=2mm,
    right=2mm,
    top=1mm,
    bottom=1mm,
    fonttitle=\bfseries\small,
    width=\linewidth
}
\usepackage[normalem]{ulem} 
\newcommand{\best}[1]{\textbf{\textcolor{red}{#1}}}
\newcommand{\secondbest}[1]{\textcolor{blue}{\uline{#1}}}
\usepackage{multirow}
\definecolor{cvprblue}{rgb}{0.21,0.49,0.74}
\usepackage[pagebackref,breaklinks,colorlinks,allcolors=cvprblue]{hyperref}
\def\BibTeX{{\rm B\kern-.05em{\sc i\kern-.025em b}\kern-.08em
    T\kern-.1667em\lower.7ex\hbox{E}\kern-.125emX}}
\begin{document}

\title{Understanding Pure Textual Reasoning for Blind Image Quality Assessment}

\author{Yuan Li and Shin'ya Nishida \\Kyoto University \\
    {\tt\small li.yuan.67n@st.kyoto-u.ac.jp}
}

\maketitle

\begin{abstract}
Textual reasoning has recently been widely adopted in Blind Image Quality Assessment (BIQA). However, it remains unclear how textual information contributes to quality prediction and to what extent text can represent the score-related image contents. This work addresses these questions from an information-flow perspective by comparing existing BIQA models with three paradigms designed to learn the image–text–score relationship: Chain-of-Thought, Self-Consistency, and Autoencoder. Our experiments show that the score prediction performance of the existing model significantly drops when only textual information is used for prediction. Whereas the Chain-of-Thought paradigm introduces little improvement in BIQA performance, the Self-Consistency paradigm significantly reduces the gap between image- and text-conditioned predictions, narrowing the PLCC/SRCC difference to 0.02/0.03. The Autoencoder-like paradigm is less effective in closing the image–text gap, yet it reveals a direction for further optimization. These findings provide insights into how to improve the textual reasoning for BIQA and high-level vision tasks.
\end{abstract}

\begin{IEEEkeywords}
Blind Image Quality Assessment, Self-Supervised Learning, Multimodal Model, Interpretable System
\end{IEEEkeywords}

\section{Introduction}
\label{sec:intro}

Early research~\cite{niqe,nima,musiq,topiq,dbcnn,yang2022maniqa} in Blind Image Quality Assessment (BIQA) focused mainly on score prediction, extracting visual features and mapping them to quality scores through classification or regression. Although these models achieved reasonable accuracy, their limited ability to capture higher-level cues (e.g., semantics) restricted their interpretability and generalization. With the rise of multimodal large language models (MLLMs) ~\cite{flamingo,llava,mplug2,qwen2.5}, recent approaches~\cite{liqe,depictqa,q-instruct,q-ground,q-insight,q-ponder} have begun to incorporate textual representations into BIQA. Works such as Q-Instruct~\cite{q-instruct}, DepictQA~\cite{depictqa}, and Q-Ground~\cite{q-ground} constructed extensive text-annotated datasets, laying a foundation for multimodal BIQA. Q-insight~\cite{q-insight} and Q-Ponder~\cite{q-ponder} take a different direction by avoiding costly human annotations and instead leveraging pretrained knowledge, using reinforcement learning to optimize solely for the final quality score.

\begin{figure}[t]
    \centering
    \includegraphics[width=\linewidth]{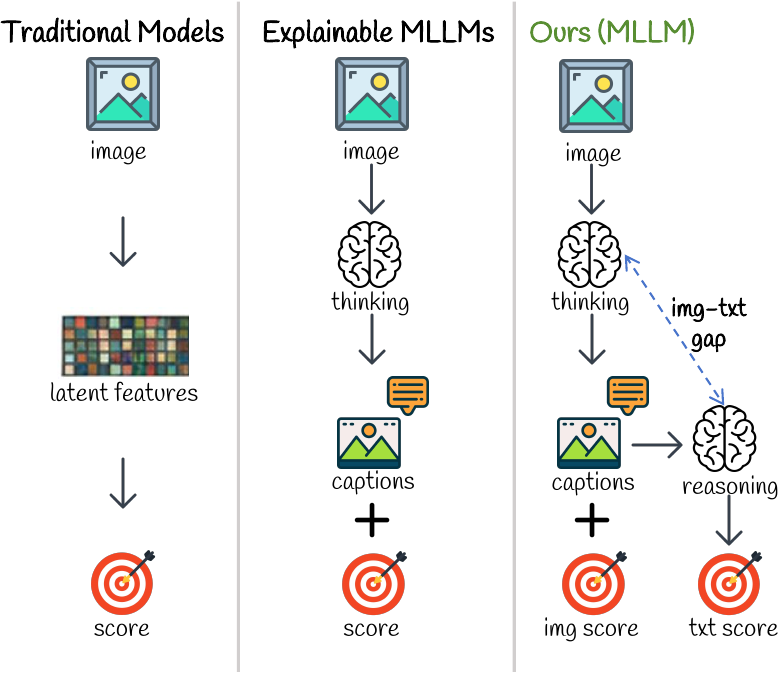} 
    \caption{ \textbf{Image-Text Gap.} Traditional BIQA relies on visual features and lacks interpretability. Although MLLM-based BIQA generates both captions and scores, their relationship remains unclear. We examine the performance gap between image-based and text-based score prediction to reveal how textual reasoning contributes to interpretable BIQA.
    }
    \label{fig_1}
    \vspace{-15pt}
\end{figure}

However, previous work on MLLM models has not clearly established the role of textural captions in BIQA apart from providing explanations. Briqa~\cite{briqa} has shown that, under supervised fine-tuning, models may bypass the intermediate text altogether when predicting scores. Q-Align~\cite{q-align} even removes the textual reasoning step entirely and directly fits MLLMs for score regression, achieving state-of-the-art (sota) performance. These observations raise an important question: how much do the generated text captions truly contribute to quality prediction, and to what extent does the model actually engage in textual reasoning rather than merely producing superficial explanations?

Motivated by these questions, we adopt an information-flow perspective to examine how effectively text alone can convey quality-related information and how different learning paradigms influence the image–text gap. To this end, we systematically study three training paradigms:
(1) a \textbf{Chain-of-Thought} paradigm,
(2) a \textbf{Self-Consistency} paradigm, and
(3) an \textbf{Autoencoder-like} paradigm.

We analyze the differences among the three paradigms from three perspectives. First, in terms of score prediction performance, the CoT paradigm provides almost no benefit, whereas both the Self-Consistency and Autoencoder-like paradigms reduce the image--text performance gap, with Self-Consistency showing the most notable improvement. Second, we examine token-level attention patterns during reasoning. The CoT paradigm behaves similarly to the baseline model, focusing on terms such as ``focus'' and ``clear.'' In contrast, the Self-Consistency paradigm shifts attention toward score-related words like ``good'' and ``moderate,'' while the Autoencoder-like paradigm highlights cues like ``blurry'' and ``focus.''  Finally, to assess whether the score-related words introduce shortcuts, we remove such score-related terms. The resulting performance drop is negligible, indicating that the models rely on additional implicit quality cues and possess stronger reasoning ability. These findings clarify how training paradigms shape internal reasoning and provide insights for developing more reliable BIQA systems.

Our contributions are two-fold:
\begin{itemize}
    \item From an information-flow perspective, we systematically evaluate three
training paradigms for learning textual representations of image quality. Our
analysis provides a structured baseline for studying textual reasoning in BIQA
and clarifies how different paradigms shape text-conditioned performance.
    
    \item Our framework is general and can be applied to other downstream tasks that
lack intermediate reasoning annotations. It offers a mechanism to induce
task-specific and interpretable textual explanations, enabling broader use in
multimodal and vision-centric applications.
\end{itemize}

\section{Related Works}
\subsection{MLLM-based BIQA Systems}

With the rapid progress of MLLMs, which demonstrate strong capabilities in textual description and reasoning, recent researches~\cite{depictqa,q-instruct,q-ground,q-insight,q-ponder} have begun to explore training MLLMs for BIQA through supervised fine-tuning (SFT) or reinforcement learning (RL). These approaches aim to construct a more interpretable assessment system. Representative works such as DepictQA~\cite{depictqa}, Q-Instruct~\cite{q-instruct}, Q-Insight~\cite{q-insight}, and Q-Ponder~\cite{q-ponder} are all devoted to building text-based interpretable systems for BIQA. Despite the presence of textual reasoning processes, these methods generally lack explicit evaluation of the quality or validity of the generated explanations. HumanIqa~\cite{humaniqa} supervises the reasoning and compares prediction performance under image- and text-conditioned settings. However, it relies on additional human-annotated reasoning data and provides limited analysis of how text-conditioned learning itself emerges or contributes to BIQA performance.

Motivated by these observations, we aim to investigate how textual tokens contribute to BIQA and how the gap between image- and text-conditioned performance can be effectively bridged, thus providing deeper insight into the role of language in MLLM-based BIQA systems.

\subsection{Interpretable Visual Reasoning}

Visual reasoning is a fundamental task of visual question answering (VQA) and MLLMs. Its primary goal is to infer answers by analyzing visual content and following a structured reasoning process. In many standard visual reasoning tasks, models benefit from abundant supervision such as question--answer pairs, attribute annotations, or annotated explanations that explicitly guide the intermediate steps of reasoning. However, for more complex perceptual tasks such as BIQA, effective supervision for intermediate reasoning is largely unavailable. To address the broader challenge of missing reasoning supervision, recent general-purpose reasoning systems have explored self-improving or self-rewarding strategies. MM-CoT~\cite{mm-cot} demonstrates how modality-specific CoT learning in both image and text domains enhances the reasoning capabilities of MLLMs, while Vision-SR1~\cite{vision-SR1} shows that such models can further improve by generating and evaluating their own reasoning processes without relying on external annotations. 

Inspired by these, we propose a related self-consistency strategy tailored for BIQA. Our method leverages the model's pre-trained image captioning capability as a form of self-supervised signal, encouraging the model to refine its internal reasoning pathway and learn a more coherent quality projection, even in the absence of explicit CoT labels.

\section{Methods}

\begin{figure}[t]
    \centering
    \includegraphics[width=\linewidth]{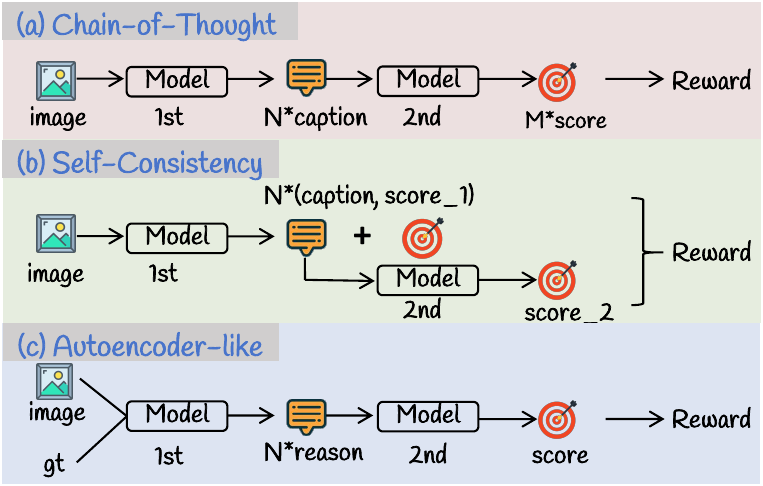} 
    \caption{\textbf{Training Paradigms.}
All models share the same MLLM backbone and perform two forward passes, where ``1st'' and ``2nd'' denote the first and second forward inferences of the same model. The 1st pass is always conditioned on the image.
\textbf{(a) Chain-of-Thought:} The model first generates $N$ caption candidates from the image; each caption then produces $M$ score predictions through an independent second-stage inference.
\textbf{(b) Self-Consistency:} The first pass outputs $N$ (caption, score) pairs, and each caption undergoes an additional inference step for score regression, receiving a self-consistency reward.
\textbf{(c) Autoencoder-like:} The model takes the image and the ground-truth MOS during the first pass to generate reasoning text; the second pass regresses the score solely from this reasoning text.
}
    
    \label{fig2}
    \vspace{-15pt}
\end{figure}

\subsection{Overview of Information-Flows}
In this work, we investigate the information flow among image, text, and score
in BIQA. We first consider a sequential formulation in which information flows
from image to reasoning and then to score, denoted as
$\boldsymbol{I \rightarrow R \rightarrow \hat{S}}$, where each stage is treated independently
in a Markov-style manner. We refer to this paradigm as \textbf{Chain-of-Thought (CoT)},
assuming that the generated text $R$ can fully represent image quality
information and directly support score prediction. 

We also introduce a \textbf{Self-Consistency} paradigm that
relaxes the strict separation between modalities. In this setting, the model
performs score prediction in two stages: the first pass follows
$\boldsymbol{I \rightarrow (R,\hat{S})}$ with visual information preserved, while
the second pass relies solely on textual reasoning, following
$\boldsymbol{R \rightarrow \hat{S}}$. This design encourages the model to acquire text-based
reasoning ability while maintaining consistency with image-conditioned
predictions.

Finally, we move beyond the forward formulation by reversing the information
flow. In this \textbf{Autoencoder-like} paradigm, the model is explicitly provided with
the ground-truth score and trained to generate explanations that justify it,
following $\boldsymbol{(I, S^{\ast}) \rightarrow R \rightarrow\hat{S}}$. This formulation explores how
quality-aware supervision shapes textual explanations.

\subsection{Chain-of-Thought Reasoning Learning}

As illustrated in Fig.~\ref{fig2} (a), the model generates $N$ reasoning traces. The $i$-th reasoning trace produces $M$ score predictions, denoted as
$s_{i,j}$ for $i = 1,\dots,N$ and $j = 1,\dots,M$.
As in~\eqref{eq1}, each score prediction $s_{i,j}$ receives a reward
$r_{i,j}$, where $x$ is the absolute difference between the predicted score and ground-truth MOS, and $t$ controls the tolerance margin.
The reward for the $i$-th reasoning trace is then obtained by averaging
the rewards of its $M$ predictions as in~\eqref{eq2}. Higher values of $R_i$ increase the generation probability of the corresponding
sentence by strengthening the loss term defined in~\eqref{eq3}. In
~\eqref{eq3}, $\mathcal{L}$ denotes the cross-entropy loss, $I$, $R$, and $S$ denote the image, reasoning (text), and quality
score, respectively, while $\alpha$ and $\beta$ are hyper-parameters that enable or disable the loss
terms associated with different training stages.
\begin{equation}
r_{i,j}=
\begin{cases}
0.5 \big( 1 + \cos(\pi x / t) \big), & \text{if } x < t, \\
0, & \text{otherwise},
\end{cases}
\label{eq1}
\end{equation}

\begin{equation}
R_i = \frac{1}{M} \sum_{j=1}^{M} r_{i,j},
\label{eq2}
\end{equation}

\begin{equation}
\mathcal{L}_{\text{total}}
= \alpha \, \mathcal{L}(I, R)
+ \beta \, \mathcal{L}(R, S^\ast).
\label{eq3}
\end{equation}

\subsection{Self-Consistency Learning}

In the first stage, the model generates a caption sequence and a score prediction directly from the image. In the second stage, the model performs another round of inference using only the generated reasoning sequence. Both predictions are supervised using the same score reward as in~\eqref{eq1}, encouraging the model to produce reasoning that is not only consistent with the visual input but also predictive when used independently as in Fig.~\ref{fig2} (b). The loss function is defined as:
\begin{equation}
\mathcal{L}_{\text{total}}
= \alpha \, \mathcal{L}(I, S^\ast)
+ \beta \, \mathcal{L}(R, S^\ast),
\label{eq4}
\end{equation}
This formulation allows the model to retain rich visual cues during training while progressively aligning its internal reasoning with textual explanations, ultimately improving its text-only reasoning capability.

\subsection{Autoencoder-like Learning}

During training stage one, the model is given the ground-truth quality score $S^\ast$ and generates a textual explanation $R$ conditioned on both the image and the score: $(I, \, S^\ast) \rightarrow R$.
In the training stage two, the model is required to perform prediction using only the generated reasoning: $R \rightarrow \hat{S}$.
The training stage two evaluates whether the explanation itself is predictive of image quality, functioning analogously to a decoding step in an autoencoder. In the testing stage, the score is masked with placeholders (e.g., “some score”). The loss from the two stages is estimated as:
\begin{equation}
\mathcal{L}_{\text{total}}
= \alpha \, \mathcal{L}(S^\ast, R)
+  \beta\,\mathcal{L}(R, S^\ast).
\label{eq5}
\end{equation}

Compared with the self-consistency paradigm, this Autoencoder-like framework explicitly places the score at the input side of reasoning generation and prohibits visual access during score regression. This encourages the model to encode score-relevant semantics into the reasoning itself, reinforcing the quality-predictive capacity of textual explanations. 

\subsection{Training via Group-Relative Policy Optimization Strategy}
To reduce the need for human annotations and enable a more scalable training paradigm, we adopt a self-supervised reinforcement learning approach built upon the Group-Relative Policy Optimization (GRPO)~\cite{guo2025deepseek} framework. During each training iteration, the model generates multiple candidate reasoning chains and corresponding answers. These candidates are then evaluated using a set of designed reward functions, which guide the optimization direction and determine how the model evolves over time. The training objective of a single GRPO process is mathematically expressed in~\eqref{grpo}.
\begin{equation}
\begin{aligned}
    \mathcal{J} _{GRPO} =  \mathbb{E}[\frac{1}{N}{\textstyle \sum_{i}^{N}}min(d_iA_i, C_{d_i,\epsilon}A_i-\beta\cdot \text{KL})],
\end{aligned}
\label{grpo}
\end{equation}
where $d_i = \frac{\pi_{\theta}(y_i|x)}{\pi_{{\theta}_{\text{old}}}(y_i|x)}$, $A_i = \frac{r_i-mean(r_1,r_2,...,r_N)}{std(r_1,r_2,...,r_N)}$, $C_{d_i,\epsilon}=clip(d_i,1-\epsilon,1+\epsilon)$, and $\text{KL} = \mathbb{D}_{\text{KL}}(\pi_{\theta}||\pi_{\text{ref}})$. Note that $\pi_{\theta}$, $\pi_\text{old}$ and $\pi_\text{ref}$ denote the policy model, old policy model and reference model, respectively. $r_i$ denotes rewards, and $\epsilon$ and $\beta$ denote hyper-parameters. The model selectively reweights different sampling losses based on the group reward, thereby reinforcing the more favorable reasoning trajectories.

\section{Experiments}
\subsection{Datasets and Training Details}
\label{sec:datasets_training}
\textbf{Training Dataset.}
To ensure a fair comparison with prior works, we adopt the default training split of the KonIQ~\cite{koniq} dataset at a resolution of $512 \times 384$. For score supervision, we use the DeQA~\cite{deqa} normalized quality labels following recent MLLM-based BIQA methods. 
\textbf{Test Datasets.}
To comprehensively evaluate generalization robustness, we test our models on six datasets, including SPAQ~\cite{spaq}, LIVE-W~\cite{live-w}, KADID~\cite{kadid}, AGIQA~\cite{agiqa}, CSIQ~\cite{csiq} and the test split of KonIQ~\cite{csiq} dataset. This mixture of real and synthetic benchmarks enables a thorough assessment of the proposed visual-to-text learning paradigms. 
\textbf{Training Details.} We use \textit{Qwen-VL-2.5-7B-Instruct~\cite{qwen2.5}} as our backbone model. In the pretraining stage, the model is optimized using only discrete score supervision together with a format reward, following Q-Insight-Score~\cite{q-insight}. We adopted the Adam optimizer~\cite{adam}, a batch size of 128, and trained for 10 epochs on eight NVIDIA A6000 GPUs, around 27 hours. 
For the main experiments, we fine-tuned each of the proposed frameworks on the same KonIQ~\cite{koniq} training split for 2 epochs, using the same configuration as in pretraining. Details are reported in Table~\ref{tab:ablation}.

\subsection{Quality Score Prediction Performance}

\begin{table*}[t]
\renewcommand{\arraystretch}{1.15}
\centering
\caption{\textbf{PLCC / SRCC performance comparisons.} 
All models are assumed to be trained on the KonIQ~\cite{koniq} training set. 
The best and second-best results are highlighted in \best{red} and \secondbest{underlined blue}. Text-Only Conditions uses generated captions; the Score-related Words Removed setting evaluates captions with terms like “good,” “moderate,” “average,” “poor,” and “decent’’ removed. Image-conditioned results of other models are from reported versions.} 
\label{tab:score}
\vspace{-5pt}

\resizebox{\textwidth}{!}{
\begin{tabular}{l|rrrrrrr}
\toprule
\textbf{Model} & \textbf{KonIQ~\cite{koniq}} & \textbf{SPAQ~\cite{spaq}} & \textbf{KADID~\cite{kadid}} & \textbf{LIVE-W~\cite{live-w}} & \textbf{AGIQA~\cite{agiqa}} & \textbf{CSIQ~\cite{csiq}} & \textbf{AVG.} \\
\midrule

\multicolumn{8}{@{\hskip 0pt}c@{\hskip 0pt}}{\textbf{\textit{Deep-Learning Models}}} \\
\midrule
NIMA~\cite{nima} (2018) & 
0.896 / 0.859 & 0.838 / 0.856 & 0.532 / 0.535 & 0.814 / 0.771 & 0.715 / 0.654 & 0.695 / 0.649 & 0.748 / 0.721 \\
DBCNN~\cite{dbcnn} (2019) & 
0.884 / 0.875 & 0.812 / 0.806 & 0.497 / 0.484 & 0.773 / 0.730 & 0.641 / 0.648 & 0.586 / 0.572 & 0.714 / 0.689 \\

MUSIQ~\cite{musiq} (2021) & 
0.924 / 0.929 & 0.868 / 0.863 & 0.575 / 0.556 & 0.789 / 0.830 & 0.722 / 0.630 & 0.771 / 0.710 & 0.775 / 0.753 \\
MANIQA~\cite{yang2022maniqa} (2022) & 
0.849 / 0.834 & 0.768 / 0.758 & 0.499 / 0.465 & 0.849 / 0.832 & 0.723 / 0.636 & 0.623 / 0.627 & 0.719 / 0.692 \\
CLIP-IQA+~\cite{wang2023exploring} (2023) & 
0.909 / 0.895 & 0.866 / 0.864 & 0.653 / 0.654 & 0.832 / 0.805 & 0.736 / 0.685 & 0.772 / 0.719 & 0.795 / 0.770 \\

\midrule
\multicolumn{8}{@{\hskip 0pt}c@{\hskip 0pt}}{\textbf{\textit{SFT-based and RL-based MLLMs}}} \\
\midrule
C2Score~\cite{zhu2024adaptive} (2024) &
0.923 / 0.910 & 0.867 / 0.860 & 0.500 / 0.453 & 0.786 / 0.772 & 0.777 / 0.671 & \secondbest{0.735 / 0.705} & 0.765 / 0.729 \\
Q-Align~\cite{q-align} (2024) &
\secondbest{0.941 / 0.940} & 0.886 / 0.887 & 0.674 / 0.684 & 0.853 / \secondbest{0.860} & 0.772 / 0.735 & 0.671 / 0.737 & 0.799 / \secondbest{0.807} \\
DeQA~\cite{deqa} (2025) &
\best{0.953 / 0.941} & \secondbest{0.895} / \secondbest{0.896} & \secondbest{0.694} / 0.687 & \best{0.892 / 0.879} & \secondbest{0.809} / 0.729 & \best{0.787 / 0.744} & \best{0.838} / \best{0.813} \\

Q-Insight-Score~\cite{q-insight} (2025) & 
0.918 / 0.895 & \best{0.903} / \best{0.899} & 
\best{0.702} / \best{0.702} & 0.870 / 0.839 &  \best{0.816} / \secondbest{0.766} &
0.685 / 0.640 &  \secondbest{0.813} / 0.789\\

\textbf{Ours (Chain-of-Thought)} &
0.920 / 0.907 & 0.886 / 0.884 & 0.629 / 0.699 & \secondbest{0.878} / 0.851 & 0.806 / \best{0.767} & 0.687 / 0.634 & 0.801 / 0.790 \\
\textbf{Ours (Self-Consistency)} &
0.917 / 0.905 & 0.883 / 0.882 & 0.632 / 0.692 & 0.874 / 0.843 & 0.805 / \secondbest{0.766} & 0.704 / 0.647 & 0.803 / 0.789 \\
\textbf{Ours (Autoencoder-like)} &
0.926 / 0.912 & 0.884 / 0.882 & 0.649 / \secondbest{0.700} & 0.873 / 0.850 & 0.810 / 0.763  & 0.683 /0.636 & 0.804 / 0.791 \\

\midrule

\multicolumn{8}{@{\hskip 0pt}c@{\hskip 0pt}}{\textbf{\textit{Text-Only Conditions}}} \\
\midrule

Q-Insight-Score~\cite{q-insight} (2025) & 
0.859 / 0.827 & 0.832 / 0.833 & 
0.604 / 0.620 & 0.778 / 0.776 &  
0.766 / 0.690 & 0.582 / 0.535 & 0.737 / 0.713  \\
\textbf{Ours (Chain-of-Thought)} &
0.851 / 0.819 & 0.829 / 0.833 & 
0.604 / 0.620 & 0.779 / 0.776 &  
0.766 / 0.690 & 0.582 / 0.535 & 0.735 / 0.712  \\
\textbf{Ours (Self-Consistency)} &
\textbf{0.900 / 0.879} & \textbf{0.864 / 0.861} & \textbf{0.627 / 0.661} & \textbf{0.838 / 0.815} & \textbf{0.797 / 0.734} & \textbf{0.672 / 0.620} & \textbf{0.783 / 0.762} \\
\textbf{Ours (Autoencoder-like)} &
0.877 / 0.861 & 0.824 / 0.839 & 
0.632 / 0.645 & 0.761 / 0.767 &  
0.774 / 0.696 & 0.585 / 0.557 & 0.742 / 0.725  \\

\midrule

\multicolumn{8}{@{\hskip 0pt}c@{\hskip 0pt}}{\textbf{\textit{Text-Only Conditions (Score-related Words Removed)}}} \\
\midrule

Q-Insight-Score~\cite{q-insight} (2025) & 
0.856 / 0.825 & 0.831 / 0.832 & 
0.611 / 0.621 & 0.772 / 0.771 &  
0.766 / 0.689 & 0.583 / 0.535 & 0.736 / 0.712  \\
\textbf{Ours (Chain-of-Thought)} &
0.851 / 0.818 & 0.831 / 0.831 & 
0.609 / 0.621 & 0.772 / 0.771 &  
0.766 / 0.689 & 0.581 / 0.534 & 0.735 / 0.711  \\
\textbf{Ours (Self-Consistency)} &
\textbf{0.898 / 0.879} & \textbf{0.861 / 0.859} & \textbf{0.632 / 0.660} & \textbf{0.829 / 0.812} & \textbf{0.794 / 0.727} & \textbf{0.665 / 0.621} & \textbf{0.780 / 0.760} \\
\textbf{Ours (Autoencoder-like)} &
0.867 / 0.847 & 0.834 / 0.838 & 
0.638 / 0.644 & 0.758 / 0.763 &  
0.774 / 0.696 & 0.589 / 0.557 & 0.743 / 0.724  \\

\bottomrule
\end{tabular}
}
\vspace{-8pt}
\end{table*}

As shown in Table~\ref{tab:score}, our models achieve performance competitive with current sota BIQA approaches across multiple benchmarks under image conditions, with the average PLCC/SRCC gap limited to 0.03/0.02 over six datasets. Remarkably, in the text-only inference setting, our Self-Consistency model reaches performance comparable to deep learning–based BIQA frameworks, indicating that they have learned genuine reasoning patterns rather than relying solely on visual features. Furthermore, the gap between image- and text-conditioned predictions is reduced to 0.02/0.03 in terms of PLCC/SRCC, marking a substantial step toward self-consistent BIQA systems.

\subsection{Ablation Studies}
\begin{table*}[t]
\renewcommand{\arraystretch}{1.15}
\centering
\caption{\textbf{Ablation studies.}
For each model, the first row shows image-conditioned results, and the second row shows text-conditioned results. Training and inference times are measured on the KonIQ~\cite{koniq} dataset.
Baseline model is the reproduced version of Q-insight-Score~\cite{q-insight}.
}
\label{tab:ablation}
\vspace{-6pt}

\resizebox{\textwidth}{!}{
\begin{tabular}{l|rrrrrr|r|rr}
\toprule
\textbf{Setting} & KonIQ & SPAQ & KADID & LIVE-W & AGIQA & CSIQ & \textbf{AVG.} & Train (hrs/epoch) & Infer (s/img) \\
\midrule

\multirow{2}{*}{Baseline} 
& 0.920 / 0.907 & 0.885 / 0.884 & 0.629 / 0.698 & 0.879 / 0.851 & 0.807 / 0.765 & 0.687 / 0.634 & 0.801 / 0.790  & \multirow{2}{*}{$\approx$2.7}  & \multirow{2}{*}{5.95 / 3.60} \\
& \textit{0.859 / 0.827} & \textit{0.832 / 0.833} & \textit{0.604 / 0.620} & \textit{0.778 / 0.776} & \textit{0.766 / 0.690} & \textit{0.582 / 0.535} & \textit{0.737 / 0.714} &  &  \\

\midrule
\multirow{2}{*}{Chain-of-Thought ($\alpha=1, \beta = 1$)} 
&0.920 / 0.907 & {0.886} / {0.884} & 0.629 / 0.699 & {0.878 / 0.851} & 0.806 / \textbf{0.767} & {0.687 / 0.634} & {0.801} / {0.790}   & \multirow{2}{*}{$\approx$5.0} & \multirow{2}{*}{6.40 / 2.40}\\
&\textit{0.851 / 0.819} & \textit{0.829 / 0.833} & 
\textit{0.604 / 0.620} & \textit{0.779 / 0.776} &  
\textit{0.766 / 0.690} & \textit{0.582 / 0.535} & \textit{0.735 / 0.712} &  &  \\

\midrule
\multirow{2}{*}{Self-Consistency ($\alpha=0, \beta = 1$)} 
& 0.922 / 0.906 & 0.886 / 0.883 & 0.642 / 0.707 & \textbf{0.880 / 0.852} & 0.809 / \textbf{0.767} & 0.700 / 0.642 & 0.807 / \textbf{0.793}  & \multirow{2}{*}{$\approx$2.6} & \multirow{2}{*}{6.07 / 3.37}\\
& \textit{0.849 / 0.810} & \textit{0.800 / 0.823} & \textit{ 0.610 / 0.642} & \textit{0.762 / 0.783} & \textit{0.769 / 0.700} & \textit{0.568 / 0.559} & \textit{0.726 / 0.720} &  &  \\

\midrule
\multirow{2}{*}{Self-Consistency ($\alpha=1, \beta = 0$)} 
& 0.917 / 0.905 & 0.883 / 0.882 & 0.632 / 0.692 & 0.874 / 0.843 & 0.805 / 0.766 & \textbf{0.704 / 0.647} & 0.803 / 0.789  & \multirow{2}{*}{$\approx$2.5} & \multirow{2}{*}{5.76 / 3.20}\\
& \textit{0.900 / 0.879} & \textit{\textbf{0.864 / 0.861}} & \textit{ 0.627 / \textbf{0.661}} & \textit{\textbf{0.838 / 0.815}} & \textit{\textbf{0.797 / 0.734}} & \textit{\textbf{0.672 / 0.620}} & \textit{\textbf{0.783 / 0.762}} &  &  \\

\midrule
\multirow{2}{*}{Self-Consistency ($\alpha=1, \beta = 1$)} 
& 0.919 / 0.907 & 0.883 / 0.883 & 0.631 / 0.700 & 0.879 / 0.849 & 0.804 / 0.766 & 0.695 / 0.631 & 0.802 / 0.789  & \multirow{2}{*}{$\approx$3.1} & \multirow{2}{*}{5.66 / 3.14}\\
& \textit{\textbf{0.881} / 0.848} & \textit{0.854 / 0.853} & \textit{ 0.621 / 0.653} & \textit{0.812 / 0.79} & \textit{0.784 / 0.704} & \textit{0.634 / 0.576} & \textit{0.764 / 0.738} &  &  \\

\midrule
\multirow{2}{*}{Autoencoder-like ($\alpha=0, \beta = 1$)} 
& \textbf{0.926 / 0.912} & 0.884 / 0.882 & \textbf{0.649} / 0.700 & 0.873 / 0.850 & 0.810 / 0.763 & 0.683 / 0.636 & \textbf{0.804} / 0.791  & \multirow{2}{*}{$\approx$7.0} & \multirow{2}{*}{6.40 / 3.52}\\
& \textit{0.877 / \textbf{0.861}} & \textit{0.824 / 0.839} & \textit{ \textbf{0.632} / 0.645} & \textit{0.761 / 0.767} & \textit{0.774 / 0.696} & \textit{0.585 / 0.557} & \textit{0.742 / 0.725} &  &  \\

\midrule
\multirow{2}{*}{Autoencoder-like ($\alpha=1, \beta = 0$)} 
& 0.919 / 0.907 & 0.885 / 0.883 & 0.618 / 0.695 & 0.879 / 0.849 & 0.804 / 0.766 & 0.691 / 0.631 & 0.799 / 0.787  & \multirow{2}{*}{$\approx$3.5} & \multirow{2}{*}{5.86 / 3.42}\\
& \textit{0.868 / 0.838} & \textit{0.838 / 0.840} & \textit{ 0.588 / 0.630} & \textit{0.803 / 0.777} & \textit{0.764 / 0.680} & \textit{0.616 / 0.548} & \textit{0.746 / 0.719} &  &  \\

\midrule
\multirow{2}{*}{Autoencoder-like ($\alpha=1, \beta = 1$)} 
& 0.925 / 0.908 & \textbf{0.887 / 0.885} & 0.648 / \textbf{0.701} & 0.876 / 0.851 & \textbf{0.812} / 0.763 & 0.695 / 0.645 & 0.799 / 0.792  & \multirow{2}{*}{$\approx$5.5} & \multirow{2}{*}{6.06 / 3.54}\\
& \textit{0.876 / 0.853} & \textit{0.827 / 0.837} & \textit{ 0.614 / 0.631} & \textit{0.767 / 0.787} & \textit{0.774 / 0.689} & \textit{0.601 / 0.587} & \textit{0.743 / 0.731} &  &  \\

\bottomrule
\end{tabular}
}
\vspace{-8pt}
\end{table*}
We conduct ablation experiments by varying the loss weights $\alpha$ and $\beta$ to control the contribution of each inference stage, as summarized in Table~\ref{tab:ablation}. We observe that the Chain-of-Thought does not gain any improvement. It may be related to that images typically contain richer and more fine-grained cues than text, and completely removing visual signals during score regression often makes it difficult to learn effective reasoning. Compared to others, the Self-Consistency model achieves the best text-only inference performance when $\alpha=1$ and $\beta=0$, whereas the Autoencoder-like model performs best under the combined image- and text-conditioned settings when $\alpha=0$ and $\beta=1$. This indicates that Self-Consistency learns the text-to-score mapping primarily when visual information is available during training, while the Autoencoder-like paradigm enhances both image-conditioned and text-conditioned performance by explicitly learning the reasoning-to-score relationship.

\begin{figure*}[t]
    \centering
    \includegraphics[width=\linewidth]{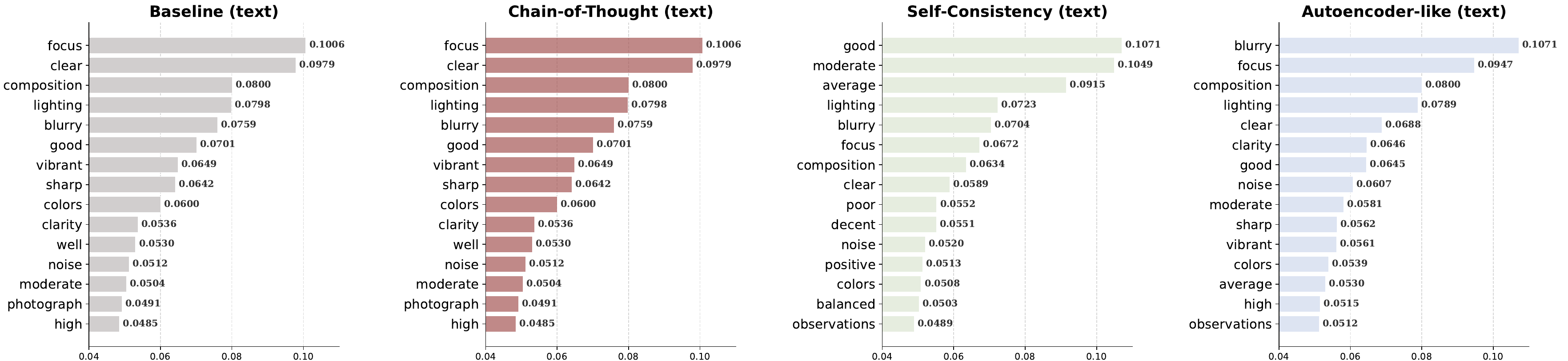} 
    \caption{\textbf{Comprehensive Analysis of Attention Distributions.}
    Please zoom in to check the details. We analyze model behavior on the KonIQ~\cite{koniq} test set (3{,}015 images) by examining softmax-normalized attention values to estimate token contributions to score
prediction. This shows the text-conditioned results. Different learning paradigms lead to
distinct shifts in the tokens emphasized by the model. Image-conditioned results are shown in Section 3.2 in supplemental materials.
}
\vspace{-15pt}
    \label{fig3}
\end{figure*}

\section{Discussion}
\subsection{Image–Text Gap}
To better understand the performance differences between image- and text-conditioned inferences, we further analyze where the score-related information originates in Fig.~\ref{fig3}. Under the image-conditioned setting (details in supplemental materials), although the softmax-normalized attention weights of non-image tokens are around 0.06, their contributions remain negligible because the unscaled attention values of image tokens are substantially larger than those of other tokens. As a result, score prediction is dominated almost entirely by image tokens. In contrast, under the text-conditioned setting, the absence of dominant image-token activations allows other tokens to contribute more effectively to the prediction.

Compared with the baseline, the Chain-of-Thought model shows a negligible change in token usage. The Self-Consistency model, however, focuses on score-related tokens such as “good,” “moderate,” and “average,” which potentially explains its strong text-conditioned performance. By contrast, the Autoencoder-like model focuses on more natural quality cues, including “blurry,” “focus,” and “composition.” This behavior enables it to improve both image-conditioned and text-conditioned performance. In the ``Score-related Words Removed'' setting of Table.~\ref{tab:score}, models remain capable of producing reasonable predictions even after removing terms such as ``good,'' ``moderate,'' ``average,'' ``poor,'' and ``decent.'' This indicates that the model's reasoning ability has improved and that it remains robust without relying on these superficial cues.

\section{Conclusion}
In this work, we investigated the information flow among image, text, and score
in Blind Image Quality Assessment. By systematically comparing three learning
paradigms—Chain-of-Thought, Self-Consistency, and an Autoencoder paradigm—we
analyzed how textual reasoning contributes to quality prediction and how the
image-text performance gap can be reduced. Our results show that naive
Chain-of-Thought reasoning has a limited impact, while Self-Consistency and
Autoencoder-like paradigms improve text-conditioned BIQA through distinct
mechanisms. In particular, Self-Consistency effectively narrows the image--text
gap, whereas the Autoencoder-like paradigm promotes more natural, quality-related
textual explanations. Through token-level analysis, we further revealed how
different training strategies shape the model’s reasoning focus. Overall, this
study provides insights into the role of textual reasoning in BIQA and offers a
principled basis for developing more interpretable quality
assessment systems. We hope these insights inspire future work on integrating
perceptual cues with textual explanations.

\clearpage
\bibliographystyle{IEEEbib}
\bibliography{icme2025references}

\vspace{12pt}

\end{document}